# Discovering Governing Equations by Machine Learning implemented with Invariance

Chao Chen[1,2,3], Xiaowei Jin[1,2,3] Hui Li[1,2,3*]

[1] Key Lab of Smart Prevention and Mitigation of Civil Engineering Disasters of the Ministry of Industry and Information Technology, Harbin Institute of Technology, Harbin, 150090, China

[2] Key Lab of Structures Dynamics Behavior and Control of the Ministry of Education, Harbin Institute of Technology, Harbin, 150090, China

[3] School of Civil Engineering, Harbin Institute of Technology, Harbin, 150090, China

[*]Corresponding author: Prof. Hui Li, lihui@hit.edu.cn

## ABSTRACT

The partial differential equation (PDE) plays a significantly important role in many fields of science and engineering. The conventional case of the derivation of PDE mainly relies on first principles and empirical observation. However, the development of machine learning technology allows us to mine potential control equations from the massive amounts of stored data in a fresh way. Although there has been considerable progress in the data-driven discovery of PDE, the extant literature mostly focuses on the improvements of discovery methods, without substantial breakthroughs in the discovery process itself, including the principles for the construction of candidates and how to incorporate physical priors. In this paper, through rigorous derivation of formulas, novel physically enhanced machining learning discovery methods for control equations: GSNN (Galileo Symbolic Neural Network) and LSNN (Lorentz Symbolic Neural Network) are firstly proposed based on Galileo invariance and Lorentz invariance respectively, setting forth guidelines for building the candidates of discovering equations. The adoption of mandatory embedding of physical constraints is fundamentally different from PINN in the form of the loss function, thus ensuring that the designed Neural Network strictly obeys the physical prior of invariance and enhancing the interpretability of the network. By comparing the results with PDE-NET in numerical experiments of Burgers equation and Sine-Gordon equation, it shows that the method presented in this study has better accuracy, parsimony, and interpretability.

# 1 INTRODUCTION

The partial differential equation (PDE) is of great significance in many areas of science and engineering, such as Navier-Stokes equations in fluid mechanics[1], Maxwell's equations in electromagnetism[2], and Schrödinger's equations in quantum mechanics[3]. Traditionally, the development of PDE depends on first principles and empirical observation, but it is difficult to use first principles or empirical observation in many complex situations, such as non-Newtonian flow, dilute flow, multiphase flow, neuroscience, bioscience, etc. The theoretical study and modelling of these complicated nonlinear dynamical systems still need to be perfected. In the past few decades, due to the advances in computer data processing capabilities, the data-driven discovery of PDE has become an emerging research hotspot.

Data-driven modelling of nonlinear dynamical systems has been the focus of research in various complex fields for decades[4-6]. Current machine learning approaches for data-driven discovery of PDE can be divided into two categories according to the optimization algorithms. The first class utilizes sparse constrained optimization techniques. Sparse identification of control equations for nonlinear dynamical systems was initially introduced by Bruntun et al. [7-9] and has been successfully applied in the situation of ordinary differential equations (ODE), including the Lorentz system as well as paradigmatic, bifurcation, and parametric systems, etc. Rudy et al. [10] presented a sequential threshold ridge regression (STRidge) methodology capable of discovering PDE from spatio-temporal measurement datasets. Schaeffe [11] employed L1 regularized least squares to accurately select active terms from the candidates and recognize the correlation coefficients, further validating the effectiveness of sparse regression. Zhang and Ma [12] combined molecular simulation with sparse regression to exactly find the governing equations of dilute gases, broadening the application scenario of data-driven discovery of PDE. Chang and Zhang [13-14] put forward a combination of data-driven and data assimilation techniques, which can infer the uncertain parameters of the nonlinear model while discovering the governing equations, moreover, the authors found subsurface flow equations from data via machine learning technique-LASSO.

Due to sparse data with high noise will bring a great difficulty for calculating the derivatives, Xu et al. [15] used the automatic differentiation of Neural Network to compute the derivatives and then conducted the PDE recognition by sparse regression which achieved satisfactory results, Chen et al. [16] proceeded to verify the efficacy of this strategy by physical informed Neural Network. Practically, the coefficients of partial differential equations are usually not a constant, Rudy et al. [17] tackled this problem with group sparsity which can get the active terms from the candidates alongside the time or space dependence of the coefficients. Jiang et al. [18] made a detailed analysis of the results by applying PDE-FIND and LASSO to actual flow fields respectively, including two-dimensional cylindrical winding flow, top-cap driven square cavity flow, and three-dimensional channel turbulence. Hu et al. [19] found that the Bayesian sparse identification with Laplace prior can work well to restore the sparse coefficients of time-varying partial differential equations. Zhang and Liu [20] intensified the processing of data with large uncertainty by asymptotic sparse technology. Goyal and Benner [21] blended machine learning and Runge-Kutta-inspired dictionary-based learning to distill the control equations from the data. Sun et al. [22] leveraged splines to interpolate locally the dynamics and then employs the physical residual of sparse regression in turn to inform the spline learning.

Long et al. [23-24] advanced the convolutional symbolic Neural Network: PDE-NET which increases the capacity of the candidate in comparison with PDE-FIND and the accuracy of the coefficients of active terms, however, the results of Neural Network tend to lack parsimony. Rao et al. [25] strengthened the precision and robustness of learning spatio-temporal dynamical systems with Neural Network by embedding known boundary conditions and assumed that a certain term must exist. So et al. [26] extended the ability of PDE-NET to extract PDE from sparse data with noise through the differential spectral normalization (DSN) regularization scheme. In the beginning, Raissi and Karniadakis [27] learned the coefficients of the equation whose form is known through the Gaussian process, while the Gaussian prior hypothesis would constrain the representability of the model and cause robustness problems. Subsequently, Raissi et al. [28-30] overcome these drawbacks in the way of Neural Network-PINN which yielded promising results.

Based on genetic algorithm is the second option. All of the above methods require a complete candidate, and if the candidate is uncomplete, the PDE cannot be found. To avoid this shortcoming, an attempt made by Xu et al. [31] was to integrate Neural Networks with genetic algorithms. The automatic differentiation of Neural Network could calculate the derivatives, then the mutation and crossover of genetic algorithm could generate the active term without the necessity of building a complete candidate. To discern the form of variable coefficients, Xu et al. [32] continued to implement a stepwise adjustment strategy on top of genetic algorithms to reveal the PDE with varying coefficients. Zeng et al. [33] executes this algorithm into the actual unknown scenario to discover the macroscopic governing equation of viscous gravity current.

Despite the advancements in the data-driven discovery of PDE, there is no fundamental breakdown of the discovery process itself, including the established guidelines of the candidates and how to accommodate existing physical priors, for instance, in Newtonian mechanics, the physical law must meet the need of Galilean invariance and in Special Theory of Relativity, Maxwell's equations, etc. must satisfy Lorentz invariance. In this work, we introduce new physically enhanced machine learning discovery methods for PDE based on Galilean invariance and Lorentz invariance: GSNN and LSNN, providing criteria for the construction of candidates for the first time. The employment of mandatory embedded physical constraints is essentially different from PINN in the form of loss functions, thus guaranteeing that the Symbolic Neural Network strictly respects the given physical prior of invariance and augments the interpretability of the network.

## 2 METHODOLOGY

### 2.1 Galileo Invariance

In Newtonian physics, Galileo invariance is a fundamental physical property, that is, the governing equations are covariant and their mathematical form is invariant with respect to (w.r.t.) Galilean transformation. The Galileo invariance should be naturally satisfied when discovering governing differential equations from a massive amount of data as mentioned in Introduction.

The generalized form of differential equations in Newtonian mechanism can be written as follows

$$u_t = N\left(u, u_x, u_{xx}, u u_x, u^2 u_{xx}, \cdots; \vartheta\right) \tag{1}$$

where $u(t,\cdot): \Omega \mapsto \mathbb{R}^d$, $N(u, u_x, u_{xx}, uu_x, u^2 u_{xx}, \cdots; \vartheta) \in \mathbb{R}^d$ which is unknown and need to be discovered from given dataset, the $\vartheta$ denotes the parameters in $N(\cdot)$ and the subscript *t* and *x* denote the partial derivation with respect to time and space. Our goal is to discover the analytic form of governing equations from the given datasets $u(t,x)$ over a certain temporal and spatial domain, $\{u(t,x): t \in \mathbb{R}, x \in \Omega \subset \mathbb{R}\}$.

The Symbolic Neural Network (SNN) [34-35] has been extensively employed to model $N$ in Eq. (1), as shown in Fig.1, which is denoted by $\mathcal{P}^k(u, u_x, u_{xx}, \cdots)$. The degree of monomials in the $\mathcal{P}^k(u, u_x, u_{xx}, \cdots)$ is no more than $k$ which is identical with the depth of the network[24] and the candidate terms generated by SNN as follows

$$\mathcal{P}^k(u, u_x, u_{xx}, \cdots) = \begin{bmatrix} 1, & \cdots, & u^k, & u_x, & \cdots, & u^{k-1}u_x, & u_{xx}, & \cdots, & u^{k-1}u_{xx}, & \cdots \end{bmatrix} \quad (2)$$

The $(u, u_x, u_{xx}, \cdots)$ is selected as input variables of network.

Considering that both coordinate systems ($E$ and $\bar{E}$) are in inertial frames and the coordinate system $\bar{E}$ moves at a constant velocity $c$ relatively to coordinate system $E$. The Galileo coordinate transformation and velocity transformation between the two coordinate systems are written as follows

$$\bar{x} = x - ct \,; \bar{t} = t \quad (3)$$

$$\bar{u}(\bar{x}, \bar{t}) = u(x,t) - c \quad (4)$$

Based on the transformation in Eq. (3) and (4), the Eq. (1) can be derived as,

$$u_t + c u_x = N\left(u - c, u_x, u_{xx}, (u-c)u_x, (u-c)^2 u_{xx}, \cdots; \vartheta\right) \quad (5)$$

Since the term $u_t$ at the left hand of the Eq. (1) produces a $cu_x$ term in which $x \in \Omega \subset \mathbb{R}$ after the Galileo transformation, considering Galileo invariance, a $-(u-c)u_x$ term must appear at the right-hand side, so that it could offset the $cu_x$. After the Galileo transformation, $\mathcal{P}^k(u, u_x, u_{xx}, \cdots)$ have the following format

$$\bar{\mathcal{P}}^k\left(u-c, u_x, u_{xx}, \cdots\right)=\left[1, \quad \cdots, \quad (u-c)^k, \quad u_x, \quad \cdots, \quad (u-c)^{k-1}u_x, \quad \cdots, \quad u_{xx}, \quad \cdots\right] \quad (6)$$

where $\bar{\mathcal{P}}^k\left(u-c, u_x, u_{xx}, \cdots\right)$ denotes the candidate terms after the transformation. By comparing the candidates $\mathcal{P}^k\left(u, u_x, u_{xx}, \cdots\right)$ and $\bar{\mathcal{P}}^k\left(u-c, u_x, u_{xx}, \cdots\right)$, we find the interesting point of is that only partial derivative terms (e.g. $u_x, u_{xx}$) are meeting the requirement of Galileo invariance which can appear in the candidates, while except for term $uu_x$, other rest terms containing $u$ (e.g. $u^2 u_x$, $u^3 u_x$) cannot be included in the candidates. Therefore, we can remove the terms in $\mathcal{P}^k\left(u, u_x, u_{xx}, \cdots\right)$ that are not satisfy invariance and rebuild $\tilde{\mathcal{P}}^k\left(u, u_x, u_{xx}, \cdots\right)$ with terms meeting the requirement of invariance as follows,

$$\tilde{\mathcal{P}}^k\left(u, u_x, u_{xx}, \cdots\right)=\left[uu_x \quad u_x \quad u_{xx} \quad u_{xxx} \quad \cdots\right] \quad (7)$$

According to the discussion above, we designed a Galileo Symbolic Neural Network (GSNN), for the given datasets $\left\{\mathbf{u}(t, x, y, z) \in \mathbb{R}^n, t \in \mathbb{R}, (x, y, z) \in \mathbb{R}^3\right\}$, the multi-dimensional variable $\mathbf{u}$ can only appear in $n^2$ specific terms: $\mathbf{u} \cdot \nabla \mathbf{u}$, while terms containing multiple powers of $\mathbf{u}$, such as $\mathbf{u} \cdot \mathbf{u}$, $\mathbf{u} \times \mathbf{u}$, $\mathbf{u} \cdot \mathbf{u} \cdot \nabla \mathbf{u}$, etc., cannot be included in the candidates. For that reason, when designing the network, the variables $\mathbf{u}$ are not involved in the fully connected operation and are multiplied with the fixed terms to constitute the specific terms before reaching the last hidden layer respectively, thus the terms generated by the $GSNN_M^k$ meet the invariance requirement which is the innovation point of our network and different from network architecture used in PDE-NET. In $GSNN_M^k$, we take $M = m \cdot n$ dimensional vector $\left(\mathbf{u}, \nabla \mathbf{u}, \nabla^2 \mathbf{u}, \nabla^3 \mathbf{u}, \cdots\right) \in \mathbb{R}^M$ as input and set $k$ hidden layers, Fig.1 is the schematic diagram of the SNN architecture with two hidden layers and Table 1 is the specification of $GSNN_M^k$. The $GSNN_M^k$ can constitute all polynomials meeting the requirement of invariance of variables $\left(\mathbf{u}, \nabla \mathbf{u}, \nabla^2 \mathbf{u}, \nabla^3 \mathbf{u}, \cdots\right)$ with total number of multiplications no more than $k$.

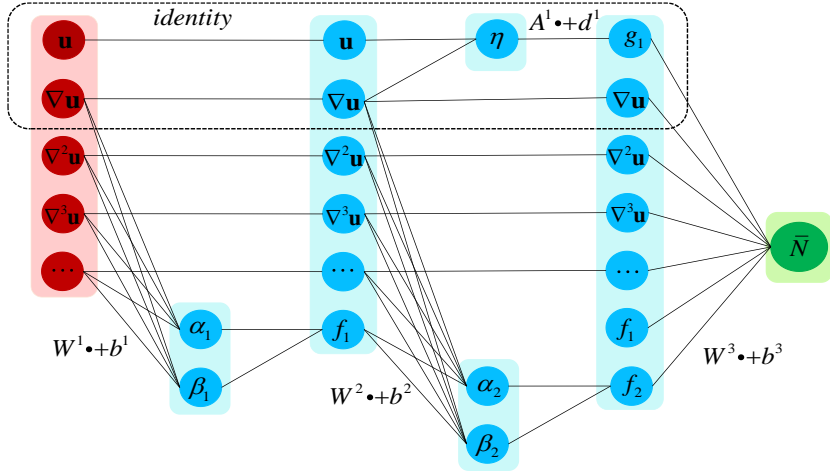

Fig. 1 The schematic diagram of the $GSNN_M^k$

Table 1 The algorithm of the $GSNN_M^k$

---

Input: $(\mathbf{u}, \nabla\mathbf{u}, \nabla^2\mathbf{u}, \nabla^3\mathbf{u}, \cdots) \in \mathbb{R}^{m \cdot n}$

$\eta = \mathbf{u} \cdot \nabla\mathbf{u}$

$g_1 = A^1 \cdot (\eta,)^T + d^1, A^1 \in \mathbb{R}^{1 \times n^2}, d^1 \in \mathbb{R}$

$(\alpha_1, \beta_1) = W^1 \cdot (\nabla\mathbf{u}, \nabla^2\mathbf{u}, \nabla^3\mathbf{u}, \cdots)^T + b^1, W^1 \in \mathbb{R}^{2 \times (m \cdot n - n)}, b^1 \in \mathbb{R}^2$

$f_1 = \alpha_1 \times \beta_1$

$(\alpha_2, \beta_2) = W^2 \cdot (\nabla\mathbf{u}, \nabla^2\mathbf{u}, \nabla^3\mathbf{u}, \cdots, f_1)^T + b^2, W^2 \in \mathbb{R}^{2 \times (m \cdot n - n + 1)}, b^2 \in \mathbb{R}^2$

$f_2 = \alpha_2 \times \beta_2$

…

$(\alpha_k, \beta_k) = W^k \cdot (\nabla\mathbf{u}, \nabla^2\mathbf{u}, \nabla^3\mathbf{u}, \cdots, f_1, \cdots, f_{k-1})^T + b^k, W^k \in \mathbb{R}^{2 \times (m \cdot n - n + k - 1)}, b^k \in \mathbb{R}^2$

$f_k = \alpha_k \times \beta_k$

Output: $\bar{N} = W^{k+1} \cdot (g_1, \nabla\mathbf{u}, \nabla^2\mathbf{u}, \nabla^3\mathbf{u}, \cdots, f_1, \cdots, f_k)^T + b^{k+1}, W^{k+1} \in \mathbb{R}^{1 \times (m \cdot n - n + k + 1)}, b^{k+1} \in \mathbb{R}$

---

Fig.2 is the schematic diagram of $\delta t$-block, we initially calculate the derivatives with the convolution operator[24]. Then in the time iteration, we applied the Euler scheme: $\tilde{\mathbf{u}}(t+\delta t,) \approx \tilde{\mathbf{u}}(t+\delta t,) + \delta t \cdot \tilde{N}$, where $\tilde{\mathbf{u}}(t+\delta t,)$ is the predicted value at time $t+\delta t$ based on the $\tilde{\mathbf{u}}(t,)$ which denotes multi-dimensional variables and $\tilde{N}$ is the multivariate nonlinear response function approximated by GSNN. In practice, we would stack multiple $\delta t$-block to improve the accuracy and anti-noise capability of the algorithm.

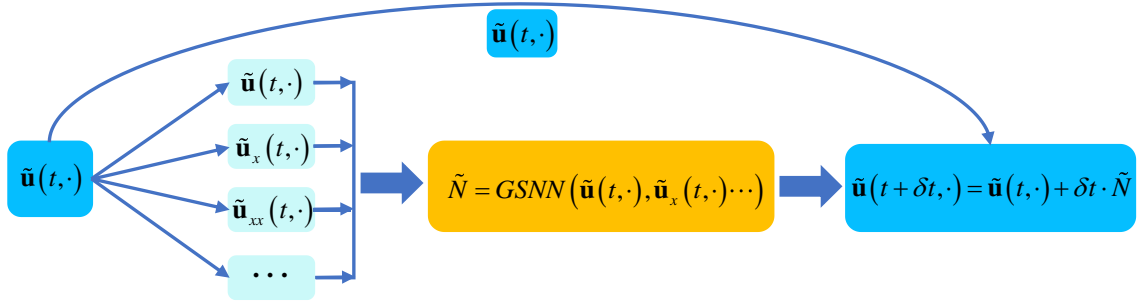

Fig. 2 The schematic diagram of the $\delta t$-block

## 2.2 Lorentz Invariance

In the context of relativity, Lorentz invariance is another essential physical property, that is governing equations are covariant and their mathematical form is invariant w.r.t. Lorentz transformation. So, when discovering the equations from a massive amount of data under the situation of relativity, we should embed the Lorentz invariance into this process.

The general form of differential equations in relativity can be written as follows:

$$u_{tt} = N\left(u, u_x, u_{xx}, uu_x, u^2 u_{xx}, \cdots; \vartheta\right) \tag{8}$$

where $u(t,\cdot):\Omega \mapsto \mathbb{R}^d$, $N\left(u, u_x, u_{xx}, uu_x, u^2 u_{xx}, \cdots; \vartheta\right) \in \mathbb{R}^d$ which is unknown and need to be discovered from the given dataset, the $\vartheta$ denotes the parameters in $N(\cdot)$ and the subscript $t$ and $x$ denote the partial derivation with respect to time and space.

We also use SNN to discover the analytic form of $N$ in Eq. (8), as shown in Fig.3, which is denoted by $\mathcal{Q}^k\left(u, u_x, u_{xx}, \cdots\right)$. The degree of monomials in the $\mathcal{Q}^k\left(u, u_x, u_{xx}, \cdots\right)$ is no more than $k$ which is identical with the depth of the network and the candidates generated by SNN as follows

$$\mathcal{Q}^k\left(u, u_x, u_{xx}, \cdots\right) = \left[1, \quad \cdots, \quad u^k, \quad u_x, \quad \cdots, \quad u^{k-1}u_x, \quad u_{xx}, \quad \cdots, \quad u^{k-1}u_{xx}, \quad \cdots\right] \tag{9}$$

The $\left(u, u_x, u_{xx}, \cdots\right)$ is selected as input variables of the network.

Considering that the coordinate system $\bar{E}$ moves at a constant velocity $c$ relatively to coordinate system $E$. The velocity $c$ is not negligible compared to the speed of light $c_0$, which is different from the Galileo transformation. The Lorentz coordinate transformation formula is

$$\bar{x} = \gamma(x - ct); \quad \bar{t} = \gamma\left(t - \frac{c}{c_0^2}x\right) \tag{10}$$

Where Lorentz factor $\gamma = \dfrac{1}{\sqrt{1 - \dfrac{c^2}{c_0^2}}}$.

Based on the transformation in Eq. (10), the Eq. (8) can be derived as,

$$\bar{u}_{\bar{t}\bar{t}} + c^2\gamma^2 \bar{u}_{\bar{x}\bar{x}} - 2c\gamma^2 \bar{u}_{\bar{x}\bar{t}} = N\left(\bar{u}, \gamma\bar{u}_{\bar{x}} - \alpha\gamma\bar{u}_{\bar{t}}, \gamma^2\bar{u}_{\bar{x}\bar{x}} + (\alpha\gamma)^2\bar{u}_{\bar{t}\bar{t}} - 2\alpha\gamma^2\bar{u}_{\bar{x}\bar{t}}, \cdots; \vartheta\right) \tag{11}$$

where $\alpha$ is $c/c_0^2$. After the Lorentz transformation the $u_x$ and $u_{xx}$ become $\gamma\bar{u}_{\bar{x}} - \alpha\gamma\bar{u}_{\bar{t}}$ and $\gamma^2\bar{u}_{\bar{x}\bar{x}} + (\alpha\gamma)^2\bar{u}_{\bar{t}\bar{t}} - 2\alpha\gamma^2\bar{u}_{\bar{x}\bar{t}}$ respectively.

Since the term $u_{tt}$ at the left hand of the Eq. (8) becomes the term $\bar{u}_{\bar{t}\bar{t}} + c^2\gamma^2\bar{u}_{\bar{x}\bar{x}} - 2c\gamma^2\bar{u}_{\bar{x}\bar{t}}$ in which $x \in \Omega \subset \mathbb{R}$ after the Lorentz transformation, considering Lorentz invariance, a term $-c_0^2\left(\gamma^2\bar{u}_{\bar{x}\bar{x}} + (\alpha\gamma)^2\bar{u}_{\bar{t}\bar{t}} - 2\alpha\gamma^2\bar{u}_{\bar{x}\bar{t}}\right)$ must appear at the right-hand side, so that it could offset the terms produced by the $u_{tt}$. After the Lorentz transformation, $\mathcal{Q}^k(u, u_x, u_{xx}, \cdots)$ has the following format:

$$\bar{\mathcal{Q}}^k\left(u, \gamma\bar{u}_{\bar{x}} - \alpha\gamma\bar{u}_{\bar{t}}, \cdots\right) = \left[1, \quad \cdots, \quad \bar{u}, \quad \gamma\bar{u}_{\bar{x}} - \alpha\gamma\bar{u}_{\bar{t}}, \quad \cdots, \quad \bar{u}(\gamma\bar{u}_{\bar{x}} - \alpha\gamma\bar{u}_{\bar{t}}), \quad \cdots\right] \tag{12}$$

where $\bar{\mathcal{Q}}^k(u, \gamma\bar{u}_{\bar{x}} - \alpha\gamma\bar{u}_{\bar{t}}, \cdots)$ denotes the candidate terms after the transformation. By comparing the candidates $\mathcal{Q}^k(u, u_x, u_{xx}, \cdots)$ and $\bar{\mathcal{Q}}^k(u, \gamma\bar{u}_{\bar{x}} - \alpha\gamma\bar{u}_{\bar{t}}, \cdots)$, we also find the interesting point is that only terms without partial derivatives (e.g. $u$, $u^2$) are meeting the requirement of Lorentz invariance which can appear in the candidates, while except for term $u_{xx}$, other rest terms containing partial derivatives (e.g. $u_x$, $uu_x$) cannot be included in the candidates. Therefore, we can remove the terms in $\mathcal{Q}^k(u, u_x, u_{xx}, \cdots)$ that are not satisfy the invariance and rebuild $\tilde{\mathcal{Q}}^k(u, u_x, u_{xx}, \cdots)$ with terms meeting the requirement of invariance as follows:

$$\tilde{\mathcal{Q}}^k(u, u_x, u_{xx}, \cdots) = \begin{bmatrix} u_{xx} & u & u^2 & u^3 & \cdots \end{bmatrix} \tag{13}$$

According to the discussion above, we designed a Lorentz Symbolic Neural Network (LSNN),

for the given datasets $\{\mathbf{u}(t,x,y,z)\in\mathbb{R}^n, t\in\mathbb{R}, (x,y,z)\in\mathbb{R}^3\}$, the partial derivatives can only appear in $n$ special terms $\nabla^2\mathbf{u}$, while other terms containing partial derivatives, such as $\nabla\mathbf{u}$, $\mathbf{u}\cdot\nabla^2\mathbf{u}$, $\nabla^3\mathbf{u}$, etc., cannot be included in the candidates. For that reason, when designing the network, the partial derivatives $\nabla^2\mathbf{u}$ are not involved in the fully connected operation and directly into the last hidden layer of network, thus the terms generated by the $LSNN_M^k$ meet the invariance requirement which is the innovation point of our network. In $LSNN_M^k$, we take $M=m\cdot n$ dimensional vector $(\nabla^2\mathbf{u},\mathbf{u},e^{\mathbf{u}},\sin\mathbf{u},\cdots)\in\mathbb{R}^M$ as input and set $k$ hidden layers, Fig.3 is schematic diagram of the SNN architecture with two hidden layers and the Table 2 is the algorithm of $LSNN_M^k$. The $LSNN_M^k$ can constitute all polynomials meeting the requirement of Lorentz invariance of variables $(\nabla^2\mathbf{u},\mathbf{u},e^{\mathbf{u}},\sin\mathbf{u},\cdots)$ with total number of multiplications no more than $k$.

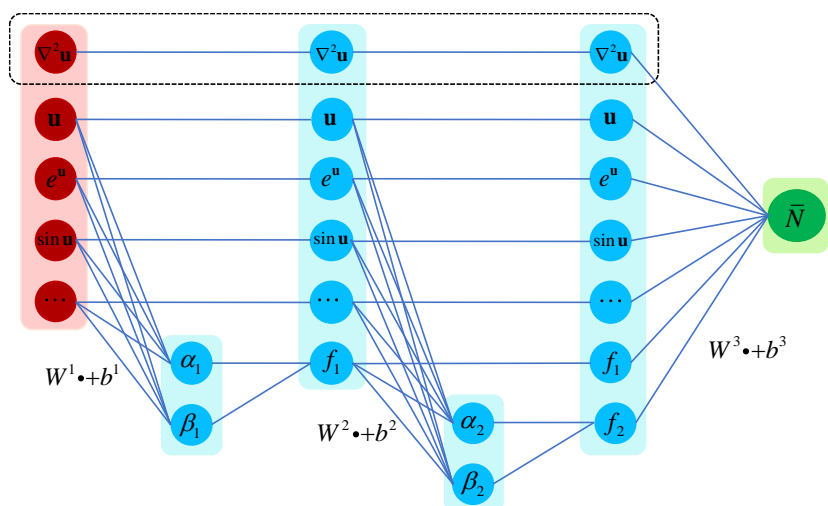

Fig. 3 The schematic diagram of the $LSNN_M^k$

Table 2 The algorithm of the $LSNN_M^k$

Input: $\left(\nabla^2 \mathbf{u}, \mathbf{u}, e^{\mathbf{u}}, \sin \mathbf{u}, \cdots\right) \in \mathbb{R}^{m \cdot n}$

$(\alpha_1, \beta_1) = W^1 \cdot \left(\mathbf{u}, e^{\mathbf{u}}, \sin \mathbf{u}, \cdots\right)^T + b^1, W^1 \in \mathbb{R}^{2 \times (m \cdot n - n)}, b^1 \in \mathbb{R}^2$

$f_1 = \alpha_1 \times \beta_1$

$(\alpha_2, \beta_2) = W^2 \cdot \left(\mathbf{u}, e^{\mathbf{u}}, \sin \mathbf{u}, \cdots, f_1\right)^T + b^2, W^2 \in \mathbb{R}^{2 \times (m \cdot n - n + 1)}, b^2 \in \mathbb{R}^2$

$f_2 = \alpha_2 \times \beta_2$

...

$(\alpha_k, \beta_k) = W^k \cdot \left(\mathbf{u}, e^{\mathbf{u}}, \sin \mathbf{u}, \cdots, f_1, \cdots, f_{k-1}\right)^T + b^k, W^k \in \mathbb{R}^{2 \times (m \cdot n - n + k - 1)}, b^k \in \mathbb{R}^2$

$f_k = \alpha_k \times \beta_k$

Output: $\bar{N} = W^{k+1} \cdot \left(\nabla^2 \mathbf{u}, \mathbf{u}, e^{\mathbf{u}}, \sin \mathbf{u}, \cdots, f_1, \cdots, f_k\right)^T + b^{k+1}, W^{k+1} \in \mathbb{R}^{1 \times (m \cdot n + k)}, b^{k+1} \in \mathbb{R}$

For the difference between the general form (1) and (8), we conceived another version of the $\delta t$-block that differs from Fig.2. In this $\delta t$-block, we added initial values $\tilde{\mathbf{u}}(-1,) = 0$ and $\tilde{\mathbf{u}}(0,) = 0$ respectively. Table 3 shows the illustration of this class $\delta t$-block. We also set multiple $\delta t$-block to increase the accuracy and anti-noise capacity of the Neural Network.

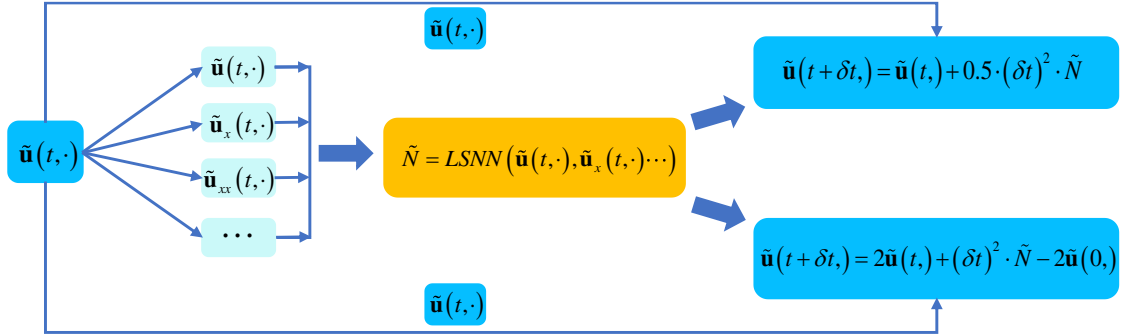

Fig. 4 The schematic diagram of the $\delta t$-block

Table 3 The illustration of $\delta t$-block

For step in blocks:
    If t = 0:
        $\tilde{\mathbf{u}}(t + \delta t,) = \tilde{\mathbf{u}}(t,) + 0.5 \cdot (\delta t)^2 \cdot \tilde{N}$
    If t > 0:
        $\tilde{\mathbf{u}}(t + \delta t,) = 2\tilde{\mathbf{u}}(t,) + (\delta t)^2 \cdot \tilde{N} - 2\tilde{\mathbf{u}}(0,)$
    $\tilde{\mathbf{u}}(0,) = \tilde{\mathbf{u}}(t + \delta t,)$

## 3 Results

In numerical experiments, we investigate the effect of considering invariance on the accuracy of active terms coefficients and the improvement of parsimony for extracting PDE with Neural Network.

## 3.1 Galileo Invariance

In this part, the Burgers equation is applied to validate the reliability of our theory. The Burgers equation is a remarkably valuable partial differential equation that can simulate the propagation and reflection of excitation waves which are common in dynamical problems including nonlinear acoustics and fluid dynamics etc.

Table 4 Discovering of Burgers equations

| Correct PDE | $u_t = -uu_x - vu_y + 0.05(u_{xx} + u_{yy})$ |
| --- | --- |
| | $v_t = -uv_x - vv_y + 0.05(v_{xx} + v_{yy})$ |
| GSNN | $u_t = -0.9820uu_x - 0.9760vu_y + 0.0465u_{xx} + 0.0454u_{yy}$ |
| | $v_t = -0.9817uv_x - 0.9770vv_y + 0.0456v_{xx} + 0.0452v_{yy}$ |
| PDE-NET | $u_t = -0.9729uu_x - 0.9717vu_y + 0.0434u_{xx} + 0.0414u_{yy}$ |
| | $v_t = -0.9778uv_x - 0.9751vv_y + 0.0430v_{xx} + 0.0424v_{yy}$ |

We begin by demonstrating the performance of GSNN in discovering the unknown PDE, and the results are presented in Table 4. As we can see from the table, GSNN has better effectiveness for viscous terms (small coefficient terms which are difficult to detect in practice), although, in convective terms (large coefficient terms), GSNN and PDE-NET have a similar effect. In addition, the terms not contained in the Burgers equation have relatively smaller weights (see Fig.5), while the number of remaining terms of GSNN is much fewer than PDE-NET and the number of remaining terms of GSNN is in a stable state, as shown in Fig.6 (here the remaining terms are those with coefficients greater than or equal to 10^-6). Those indicate that adding the physical of invariance can greatly contribute to the parsimony of discovering equations with Neural Network and the lack of parsimony is the main drawback of Neural Network method.

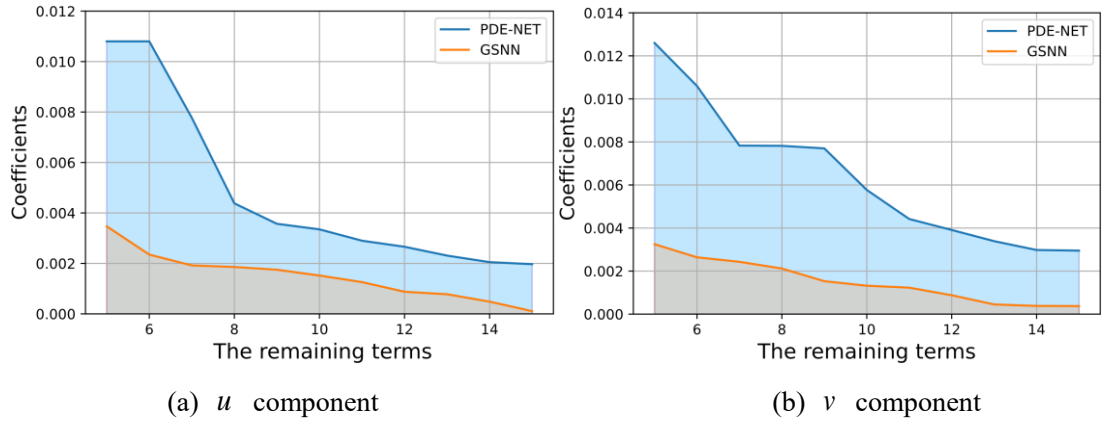

(a) $u$ component  (b) $v$ component

Fig.5 The comparison of remaining term coefficient

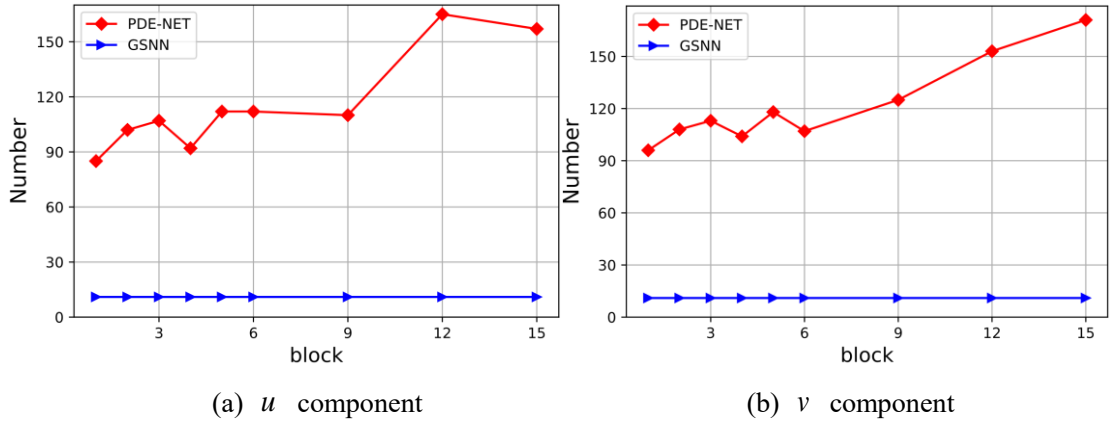

(a) $u$ component  (b) $v$ component

Fig.6 The comparison of remaining term numbers

## 3.2 Lorentz Invariance

In 1939, FRENKE and KONTOROVA studied the propagation of slip-on an infinite chain of elastically bound atoms located on a fixed atom-like lower chain. To describe this effect, they derive a differential equation that can be generalized to a partial differential equation, namely the Sine-Gordon equation[36-37]. The Sine-Gordon equation has a wide range of applications in nonlinear optics and quantum mechanics.

Table 5 Discovering of Sine-Gordon equation

| Correct PDE | $u_{tt} = 10\sin u + 0.5(u_{xx} + u_{yy})$ |
|---|---|
| LSNN | $u_{tt} = 9.8842\sin u + 0.4946 u_{xx} + 0.4617 u_{yy}$ |
| PDE-NET | $u_{tt} = 9.8356\sin u + 0.4780 u_{xx} + 0.4416 u_{yy}$ |

Again, as exhibited in table 5, LSNN has better resilience than PDE-NET which does not take the constraint of invariance into account. Meanwhile, the number of the remaining terms of LSNN is fewer than PDE-NET and the terms that are not belong to the Sine-Gordon have comparatively minor weights, as shown in Fig.7. Those also indicate that adding the physical of invariance can contribute to the parsimony of discovering equations with Neural Network

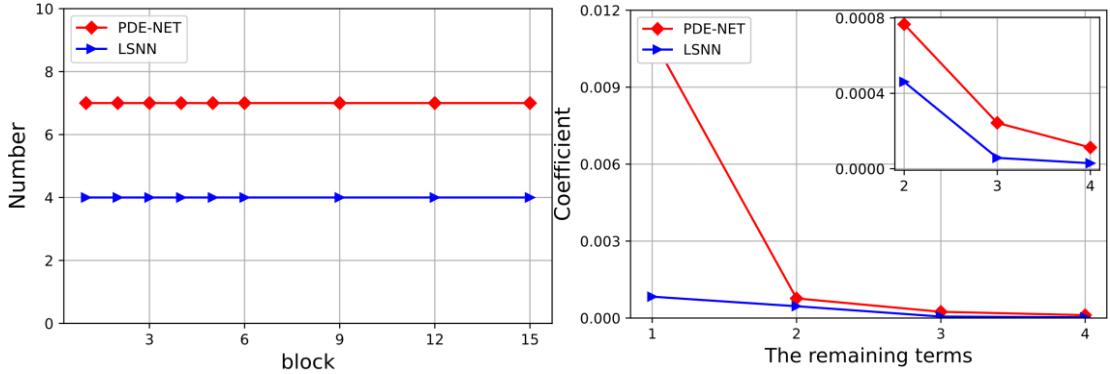

Fig. 7 The comparison of remaining term coefficient (left)

and remaining term numbers (right)

## 4 Conclusions and Future Work

In this paper, we conceive two symbolic Neural Networks for discovering PDE depending on the physical prior of invariance: GSNN and LSNN. The method of enforced embedding of physical constraint is qualitatively distinguished from PINN in the form of loss functions, which makes sure that the designed Neural Network strictly obeys the given physical prior (Galileo invariance and Lorentz invariance) and augments the explanations of Neural Network. As we can see from the numerical experiments of Burgers equation and Sine-Gordon equation that the model considering physical constraint has better accuracy than the traditional way. More importantly, taking the invariance into account can boost the parsimony of discovering PDE with Neural Network. Overall, we offer a new way for scientific discovery (extraction of PDE) and the conception of Neural Networks incorporating physical constraints. The limitations of the current work and the aspects that need further research are: (1) Since the conventional thermodynamic equations do not respect the Lorentz invariance, the physical constraint of the invariance for the modified thermodynamic equations still require study. (2) The main purpose of the methodology in this paper is for the partial differential equation. For some ordinary differential equations, say van der Pol equation, when constructing the candidates, this approach makes little sense for the reasons that the terms in the candidates are the derivatives of the coordinate $x$ and $y$ with respect to time, etc, and these terms satisfy the demands of invariance naturally. (3) At present, the forward Eulerian scheme is chosen for the time iteration, which is comparatively coarse. In the future, the Runge-Kutta scheme could be integrated with symbolic Neural Network to promote the precision and stability of the model.